\pdfoutput=1
\documentclass[11pt]{article}
\usepackage[T1]{fontenc}
\usepackage[utf8]{inputenc}
\usepackage{lmodern}
\usepackage[margin=1in]{geometry}
\usepackage{booktabs}
\usepackage{tabularx}
\usepackage{array}
\usepackage{amsmath}
\usepackage{xcolor}
\usepackage{tikz}
\usetikzlibrary{positioning,arrows.meta,calc}
\definecolor{cHum}{HTML}{E87BA4}\definecolor{cMgr}{HTML}{EB6834}\definecolor{cRes}{HTML}{2A78D6}
\definecolor{cCri}{HTML}{EDA100}\definecolor{cRev}{HTML}{1BAF7A}
\tikzset{
  pics/robot/.style={code={
    \draw[line width=1.1pt, line cap=round] (0,22) -- (0,31); \fill (0,33) circle (4);
    \fill[rounded corners=2.5pt] (-24,-16) rectangle (24,22);
    \fill (-29,-4) rectangle (-24,10); \fill (24,-4) rectangle (29,10);
    \fill[white] (-9,5) circle (5); \fill[white] (9,5) circle (5); \fill[white] (-9,-9) rectangle (9,-6);
    \fill (-16,-18) -- (16,-18) -- (20,-30) -- (-20,-30) -- cycle;}},
  pics/person/.style={code={
    \fill (0,12) circle (13);
    \fill (-25,-27) .. controls (-25,-6) and (-13,-2) .. (0,-2) .. controls (13,-2) and (25,-6) .. (25,-27) -- cycle;}},
  avatar/.style={draw=#1, fill=#1!16, line width=1.3pt},
  flow/.style={-{Stealth[length=2.2mm]}, line width=0.8pt, black!45},
  lab/.style={draw=black!22, fill=white, rounded corners=1.7mm, font=\sffamily\scriptsize, inner sep=1.6pt, minimum height=4.2mm},
  num/.style={circle, fill=black, text=white, font=\sffamily\tiny\bfseries, inner sep=0pt, minimum size=3.2mm},
  rname/.style={font=\sffamily\small\bfseries, inner sep=1pt},
  rsub/.style={font=\sffamily\scriptsize, text=black!60, inner sep=1pt}}
\newcommand{\flowlabel}[4]{\node[lab, anchor=#1] (fl) at (#2) {\hspace*{3.3mm}#4};\node[num] at ([xshift=2.1mm]fl.west) {#3};}
\usepackage{enumitem}
\usepackage[numbers,sort&compress]{natbib}
\usepackage[colorlinks=true,linkcolor=blue!50!black,citecolor=blue!50!black,urlcolor=blue!50!black]{hyperref}
\usepackage{caption}
\makeatletter\@secpenalty=0\makeatother

\newcolumntype{Y}{>{\raggedright\arraybackslash}X}


\title{Letting AI Formulate the Research Question:\\ Autonomous Research Agents}
\author{Ko Fujimura\thanks{\texttt{fujimura@otsuma.ac.jp}}\\[2pt]
{\normalsize Otsuma Women's University, Faculty of Social Information Studies}}
\date{29 October 2026}

\begin{document}
\maketitle

\begin{abstract}
Systems such as The AI Scientist have shown that AI agents can carry out and write up research, but in those demonstrations humans chose the research area and selected the outputs to show. We study letting the AI formulate the research question itself, and we report every run. Our system gives an off-the-shelf coding agent an instruction document and two skills: one for formulating the question, in which the agent diverges from recent technologies into problems of other fields, screens its candidates for known failure modes and obtains a critique from a different model; and one for peer review, in which a reviewer from another model family re-runs the study from its records before the agent revises the paper. Over 13 development runs, changes to the one or two sentences that set the research area coincided with large shifts in what the candidate questions and the chosen papers were about, which led us to the area of a new technology applied to a societal problem. We froze the system after a pilot run and report five consecutive runs: the pilot, counted by a disclosed amendment to the pre-registration, and four pre-registered runs. All five runs went from question formulation to a revised, re-computable paper, with local computation on a laptop and no research input from humans, and external AI reviewers reading only the papers judged them at or near workshop level. We release the system, the five papers and the records needed to reproduce them.
\end{abstract}

\section{Introduction}
\label{sec:intro}

Automating research end to end has moved from aspiration to demonstration. The AI Scientist \citep{lu2024aiscientist} generated machine-learning papers from code templates; The AI Scientist-v2 \citep{yamada2025aiscientistv2} removed the templates, added an agentic tree search over experiments, and had one of three generated papers reach an above-threshold score at an ICLR~2025 workshop; a later journal version describes the full pipeline and an automated reviewer \citep{lu2026nature}. Other systems automate parts of the process or target particular sciences \citep{schmidgall2025agentlab,schmidgall2025agentrxiv,gottweis2025coscientist,mitchener2025kosmos,miyai2025jraiscientist,villaescusa2025denario}, and a venue has accepted papers with AI first authors \citep{bianchi2025agents4science}.

In these demonstrations a human still decides two things. The human chooses \emph{where} to look: The AI Scientist works ``within a user-specified machine learning research subfield'' \citep{lu2026nature}, Jr.\ AI Scientist starts from a given baseline paper \citep{miyai2025jraiscientist}, and the AI co-scientist answers research goals set by scientists \citep{gottweis2025coscientist}. The human also chooses \emph{what to show}: the authors of The AI Scientist-v2 ``selected the most promising complete output'' for submission \citep{yamada2025aiscientistv2}. Within these bounds the systems generate ideas and hypotheses themselves, but the frame in which a question is formed, and the choice of which results are seen, remain with people. We therefore separate three decisions: setting the research area, formulating and selecting a specific question, and selecting which outputs to report.

We ask what happens when the area is set only broadly, the specific question is left to the AI, and nothing is filtered. The human sets the area in one or two sentences (here, societal problems that can be studied with open data); the agent writes its own candidate questions and selects one, with no human selection; and every run is reported. Our system is deliberately plain: a general-purpose coding agent, an instruction document, and two skills (structured instructions loaded on demand), with no bespoke orchestration code or experiment templates. The research-question skill is a procedure for formulating questions that targets failure modes we observed while developing the system: questions whose answer follows from definitions, technologies used as decoration, and questions with no public ground truth. A different model critiques the top candidates once, and a reviewer from another model family re-runs the study from its records before the agent revises the paper.

We froze the system after a pilot run and report five consecutive runs: the pilot and four pre-registered runs. The pilot was counted by an amendment made after three of the later results had been seen, which we report as a deviation (Section~\ref{sec:design}). Our contributions are:
\begin{enumerate}[leftmargin=*,itemsep=2pt]
  \item \textbf{Autonomy without research intervention.} All five runs completed question formulation, experiments, writing, AI review and revision, with local computation on a laptop. The records contain no research advice and no permission requests; the only inputs after the start were fixed messages to resume after usage limits and to relay the review (Section~\ref{sec:res-ops}).
  \item \textbf{Setting the research area.} Over 13 development runs, changes to the sentences of the instructions that set the area coincided with large shifts in what the candidate questions and the chosen papers were about. We describe the path to the technology $\times$ societal-problem setting and argue that more specific settings could yield correspondingly specific themes (Section~\ref{sec:area}).
  \item \textbf{A procedure for formulating research questions} and what it produced: 77 candidate questions across nine societal fields, and the reasons candidates were discarded (Section~\ref{sec:res-questions}).
  \item \textbf{Cross-model critique and review inside the loop.} The critic changed how the chosen question was compared and measured in every run, and the reviewer re-computed every first draft and found problems of interpretation that the agent then fixed (Sections~\ref{sec:res-critic}--\ref{sec:res-review}).
  \item \textbf{An external AI evaluation} of the five final papers by reviewers from two further model families (Section~\ref{sec:res-e1}), and the release of the system, the papers and the records.
\end{enumerate}

\begin{figure}[t]
\centering
\begin{tikzpicture}[scale=1.12, transform shape]
\coordinate (H) at (0,6.2); \coordinate (M) at (0,3.1); \coordinate (R) at (0,0);
\coordinate (V) at (6,3.1); \coordinate (C) at (6,0);
\draw[flow] (-0.11,5.47) -- (-0.11,3.83); \draw[flow] (0.11,3.83) -- (0.11,5.47);
\draw[flow] (-0.11,2.37) -- (-0.11,0.73); \draw[flow] (0.11,0.73) -- (0.11,2.37);
\draw[flow] (5.27,3.21) -- (0.73,3.21);   \draw[flow] (0.73,2.99) -- (5.27,2.99);
\draw[flow] (5.27,0.11) -- (0.73,0.11);   \draw[flow] (0.73,-0.11) -- (5.27,-0.11);
\draw[flow] ($(R)!0.73cm!(V)$) -- ($(V)!0.78cm!(R)$);
\draw[avatar=cHum] (H) circle (0.55); \pic[scale=0.0105] at (H) {person};
\draw[avatar=cMgr] (M) circle (0.55); \pic[scale=0.0105] at (M) {robot};
\draw[avatar=cRes] (R) circle (0.55); \pic[scale=0.0105] at (R) {robot};
\draw[avatar=cRev] (V) circle (0.55); \pic[scale=0.0105] at (V) {robot};
\draw[avatar=cCri] (C) circle (0.55); \pic[scale=0.0105] at (C) {robot};
\node[rsub] at (0,7.0) {organiser};                \node[rname] at (0,7.33) {Human};
\node[rname, anchor=east] at (-0.72,3.27) {Management agent}; \node[rsub, anchor=east] at (-0.72,2.92) {Claude Code\,·\,Opus 5.5};
\node[rname] at (0,-0.83) {Research agent};        \node[rsub] at (0,-1.16) {Codex CLI\,·\,gpt-6.1-sol};
\node[rname, anchor=west] at (6.72,3.27) {Reviewer}; \node[rsub, anchor=west] at (6.72,2.92) {Claude Code\,·\,Opus 5.5};
\node[rname, anchor=west] at (6.72,0.17) {Critic};   \node[rsub, anchor=west] at (6.72,-0.18) {API\,·\,gpt-6-astra};
\flowlabel{east}{-0.32,4.65}{2}{approval; terminal with API key}
\flowlabel{west}{0.32,5.1}{1}{instructions for the run}
\flowlabel{west}{0.32,4.65}{12}{notice of completion}
\flowlabel{west}{0.32,4.2}{13}{post-run assessment}
\flowlabel{east}{-0.32,2.0}{3}{research instructions (start)}
\flowlabel{east}{-0.32,1.55}{6}{fixed resumption messages}
\flowlabel{east}{-0.32,1.1}{10}{review report, verbatim}
\flowlabel{west}{0.32,2.15}{11}{final paper, records}
\flowlabel{center}{3.0,3.52}{9}{review report}
\flowlabel{center}{3.5,2.68}{7}{review instructions (start)}
\flowlabel{west}{3.4,1.25}{8}{draft and records (Git tag)}
\flowlabel{center}{3.0,0.42}{5}{one critique}
\flowlabel{center}{3.0,-0.42}{4}{top-5 candidates, unranked}
\end{tikzpicture}
\caption{Data flow in a run: a human and four agents (management agent, research agent, critic and reviewer). Numbers give the order within a run. The human deals only with the management agent; after the instructions, the research agent receives only fixed resumption messages, the critique it requested and the review. Steps 1 and 13 close the loop through which the instructions were revised between development runs (Section~\ref{sec:area}); during the evaluation runs the instructions were frozen.}
\label{fig:system}
\end{figure}
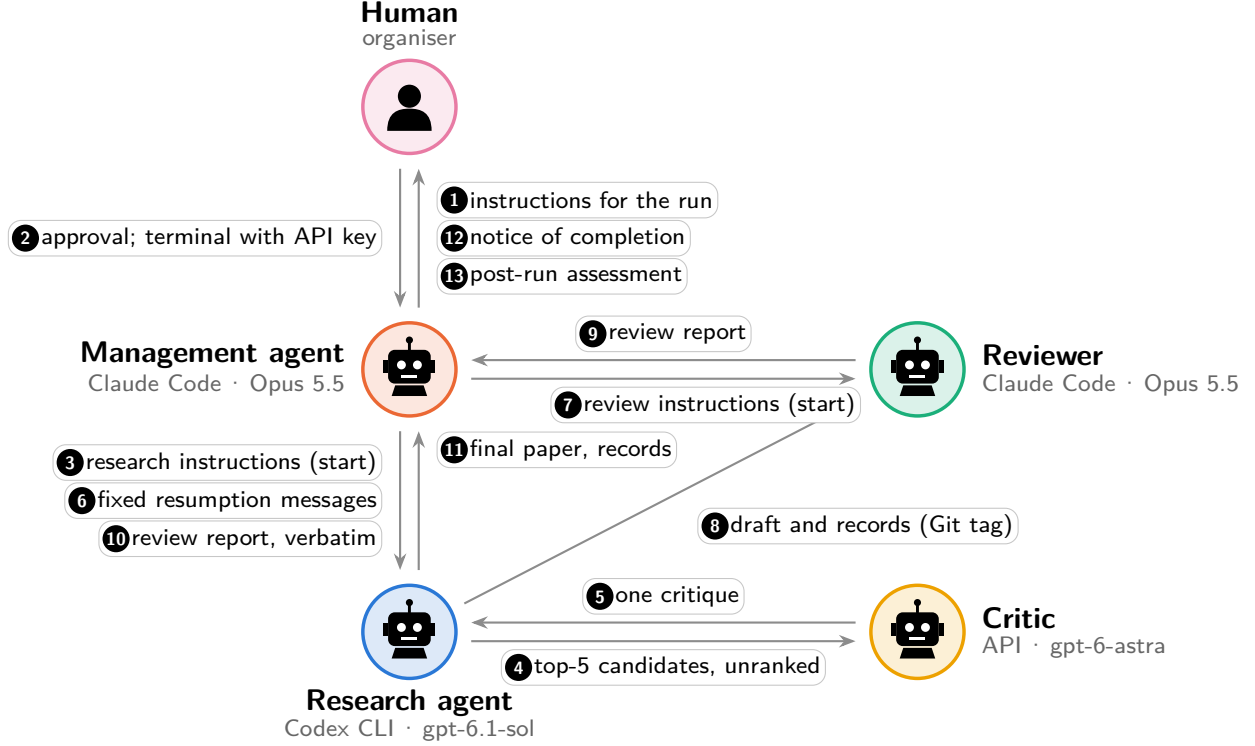

\section{Related work}
\label{sec:related}

\paragraph{End-to-end research systems.}
The AI Scientist \citep{lu2024aiscientist} introduced a pipeline of idea generation, code edits on a template, experiments, write-up and automated review. The AI Scientist-v2 \citep{yamada2025aiscientistv2} added template-free experimentation with a tree search managed by an experiment-manager agent and was evaluated by submitting three papers to a workshop; the journal version \citep{lu2026nature} reports that paper quality, as judged by its automated reviewer, improves with newer foundation models and more test-time compute. Agent Laboratory \citep{schmidgall2025agentlab} and AgentRxiv \citep{schmidgall2025agentrxiv} use LLM agents as research assistants with human input at chosen points; Jr.\ AI Scientist \citep{miyai2025jraiscientist} extends a given baseline paper and reports the risks it observed; Kosmos \citep{mitchener2025kosmos} and Denario \citep{villaescusa2025denario} target data-driven discovery; the AI co-scientist \citep{gottweis2025coscientist} generates and ranks hypotheses for scientists. Trehan and Chopra \citep{trehan2026whyllms} describe four autonomous research attempts and their failure modes. Closest to our setting, the Scientist-Bench of AI-Researcher \citep{tang2025airesearcher} includes open-ended tasks in which the agent must find its own research direction from given reference papers and datasets. Luo et al.\ \citep{luo2025pitfalls} show that failures such as inappropriate benchmark selection, data leakage, metric misuse and post-hoc selection are detected far more effectively from workflow logs and code than from the final paper alone. Our system differs in that it contains no bespoke orchestration code, sets the area only broadly instead of giving a subfield, baseline, goal or reference papers, and does not filter outputs; the records that allow re-computation are part of its product.

\paragraph{Research ideation by LLMs.}
In a large human study, LLM-generated research ideas were judged more novel but slightly less feasible than those of experts \citep{si2024ideas}; when the ideas were executed, the advantage disappeared \citep{si2025ideationgap}. An independent evaluation of The AI Scientist found weak literature review and frequent experiment failures \citep{beel2025evaluating}. A recent study asks whether an agentic workflow can propose research questions that experts in mathematics, statistics, theoretical computer science and physics find interesting \citep{nie2026after42}. We observe question formulation inside complete, unfiltered research runs, together with the critiques and reviews that followed.

\section{System}
\label{sec:system}

\subsection{Roles}
A run involves a human and four agents (Figure~\ref{fig:system}). We describe them in the order in which they act; the numbers refer to the steps in the figure.

The \textbf{human} organises the study. The human approves the instructions for each run, sets the API key in the terminal used for research and hands that terminal to the management agent (steps 1--2), and would decide any permission request; there were none in the evaluation runs.

The \textbf{management agent} is Claude Code with Claude Opus~5.5 in a separate workspace. It fills in the instructions for each run (1), starts the research session with them (3), resumes the session after usage limits with fixed messages (6), starts the reviewer once the first draft is tagged (7), and relays the review verbatim (10). After the run it reads the final paper and records (11), notifies the human (12) and writes a post-run assessment (13). It sends no research advice, and nothing it writes other than the instructions is visible to the research session. During development these assessments led to the next version of the instructions; during the evaluation runs the instructions were frozen.

The \textbf{research agent} is Codex CLI with gpt-6.1-sol (reasoning effort xhigh), started in a dedicated directory with the instruction document as its first message. It formulates the question, carries out the study and writes the paper, using its tools, web search and local computation. Through an API it may call one other model, the critic, within a budget of US\$20.

The \textbf{critic} is gpt-6-astra (reasoning effort xhigh, with web search). The research agent sends it its top five candidate questions without its scores or ranking (4) and receives one critique (5). Whether to adopt each point is the research agent's decision.

The \textbf{reviewer} is Claude Code with Claude Opus~5.5 (effort xhigh), without API keys. It works on a copy of the tagged draft and its records (8), re-runs the study through its verification entry point, checks that plans precede results, searches for prior work, and writes a review with a verdict, six scores and numbered findings (9). The research agent then has two hours to respond and submit the final version.

\subsection{Instructions and skills}
The instruction document (about 7,300 English words) states that the human will not specify the content, that negative results and failures are to be submitted as they are, and that the resource limits are for scoping a chosen question, not a source of ideas. It sets the conditions (12 hours to the draft and 2 hours for the review response, excluding usage-limit stops; 10~GiB of storage; at most two concurrent experiments of up to 30 minutes each), the records to keep, the rules on security, privacy and data about people, and a hint toward societal problems that can be studied with open data. It does not name a field or a topic.

The \textbf{research-question skill} (about 2,400 words) is the procedure for formulating the question:
\begin{enumerate}[leftmargin=*,itemsep=1pt]
  \item Record the first three candidates that come to mind as \emph{obvious candidates}; they may be adopted, but their non-obviousness score is at most 2.
  \item \emph{Diverge from technologies}: list at least six technologies from different subfields of computer science that first became possible around 2024 or later, each confirmed by a source; generalise each into capabilities without its name; for each capability, find fields outside computer science where someone could not make a judgment because of the removed constraint; check for public data; and write at least twelve questions, with no technology or field supplying more than a third of them and no type of study more than half.
  \item Check each question against six points: whether the study is possible without the capability (decoration); whether the technology or an existing method has already been applied to the same problem; whether the main result follows from definitions or how the data are produced; whether AI judgments in it can be checked against public ground truth; whether the technology or its published outputs can be used here; and which decision changes in which direction for each possible answer. A candidate that fails one of the first four checks is revised or discarded; one that fails the fifth or sixth is discarded.
  \item Check prior work for the top five, run short feasibility probes for the top three, and score every candidate from 1 to 5 on seven axes (impact, novelty, non-obviousness, generality, unpredictability, feasibility, strength of verification), with a one-line reason for each.
  \item Obtain one critique of the top five from the critic, sending only the questions, method outlines and closest prior work (not the scores or the preferred candidate); verify any prior work it cites; classify each point as adopted or not, with reasons; and re-score if needed, keeping the original scores. Rankings or new topics proposed by the critic may not be used as grounds for adopting or rejecting a candidate.
  \item Select by the total of impact, novelty, non-obviousness, generality and unpredictability; a candidate with feasibility 2 or less is first narrowed and re-scored. Write a pre-mortem (the one-sentence claim if it succeeds, the three likely first objections and the responses), and run a safety check.
\end{enumerate}
The skill also lists types of study to avoid, such as quantitative re-confirmation of a known concern and applying an existing measurement to a new target only. The checks and scores are the agent's own judgments, recorded with reasons; nothing enforces them.

The \textbf{peer-review skill} (about 1,800 words) tells the reviewer to record its prior expectation from the title and research question, to re-run the verification steps on a copy of the tagged commit without calling AI APIs, to check the timeline of plans and results, the evidence for each claim, the cited and the missing prior work, and to write the report. A separate review instruction (about 1,600 words) starts the reviewer.

\subsection{Run records and operation}
The research agent keeps the records of its run in its directory, under version control. They are append-only and consist of:
\begin{itemize}[leftmargin=*,itemsep=0pt]
  \item plans and topic-selection records, committed before the experiments that use them;
  \item a timeline of events, and every input from the human or the management agent, by type;
  \item each claim with its evidence and whether it is exploratory;
  \item each API call with its request, response and cost;
  \item an environment manifest with the deadlines;
  \item Git tags for the first draft and the final version, and a SHA-256 list of the files;
  \item a single verification entry point that regenerates the reported numbers, tables and figures without AI calls.
\end{itemize}
These records are what the reviewer checks (step 8) and what the management agent reads after the run (step 11). Research sessions use a configuration directory separate from the human's, so that the human's own skills, memories and earlier conversations are not visible; the research workspace contains no evaluations of earlier runs, and the instructions forbid reading earlier run directories, which were not blocked by access control. Usage-limit stops of the subscription-based agent are excluded from the time budget. All local computation ran on one MacBook Pro (M3 Pro, 18~GiB of memory, no discrete GPU); the language models were used as hosted services.

\section{Setting the research area: lessons from 13 development runs}
\label{sec:area}

\begin{table}[tp]
\centering
\caption{Development runs (Japanese instructions). ``Area setting'' is what the version of the instructions said about where to look; ``unchanged'' refers to this part only, since other parts of the instructions changed in every version. Candidates are counted by our classification of what each candidate studies: AI, the behaviour of AI models; Tools, tools and infrastructure for AI agents; Pract., research practice; Sys., computer systems; Math, mathematics; Soc., societal problems. The class of the chosen study is in bold.}
\label{tab:dev}
\footnotesize
\setlength{\tabcolsep}{3pt}
\begin{tabularx}{\textwidth}{@{}lllYrrrrrrY@{}}
\toprule
Run & Ver. & Model & Area setting & AI & Tools & Pract. & Sys. & Math & Soc. & Chosen study \\
\midrule
001 & v1.0 & gpt-6-sol & None: no field, no selection procedure & 0 & 0 & 0 & 0 & \textbf{2} & 0 & Non-uniform birthday problem (exact computation) \\
002 & v1.0 & gpt-6.1-sol & Field: computer science (CS) & 0 & 0 & 0 & \textbf{2} & 0 & 0 & Admission control for LRU caches \\
003 & v1.1 & Opus 5.5 & CS; at least 8 candidates; aim at questions that change someone's judgment & 2 & 1 & 1 & 3 & 0 & \textbf{3} & Third-party tracking on Japanese local-government websites \\
004 & v1.2 & Opus 5.5 & CS; ``AI-era problems'' welcomed and scored & 5 & \textbf{4} & 4 & 1 & 0 & 2 & Silent mis-application in coding-agent edit tools \\
005 & v1.2.1 & GPT-6 & unchanged (research model changed) & 0 & \textbf{8} & 2 & 1 & 0 & 0 & Cost of structured-generation engines under JSON-Schema rewrites \\
006 & v1.3 & Opus 5.5 & CS; readers are future autonomous research agents; AI-era axis removed & \textbf{9} & 2 & 7 & 1 & 0 & 0 & Do hypothesis-supporting results make LLM agents check less? \\
007 & v1.4 & Opus 5.5 & unchanged & 8 & 0 & \textbf{7} & 2 & 0 & 1 & Granularity checks of benchmark values reported in LLM papers \\
008 & v1.5 & gpt-6.1-sol & unchanged & 1 & 0 & \textbf{3} & 6 & 5 & 0 & Reuse conditions for learning-curve archives \\
009 & v1.6 & gpt-6.1-sol & CS; readers are researchers and practitioners of the field & 0 & 0 & 0 & \textbf{11} & 4 & 0 & Re-ranking to parent documents in retrieval \\
010 & v1.7 & gpt-6.1-sol & unchanged & 1 & 0 & 2 & \textbf{9} & 3 & 0 & Workflow-cache savings and postdominator bounds \\
011 & v1.8 & gpt-6.1-sol & No field; hint: discover, or help solve, societal problems using open data & 0 & 0 & 0 & 0 & 1 & \textbf{14} & Simulated re-measurement of PFAS non-detects in water monitoring \\
012 & v1.9 & gpt-6.1-sol & As 011; diverge from technologies practical since 2024 into fields outside CS & 1 & 0 & 1 & 0 & 0 & \textbf{15} & Forecasters and data recency in wastewater surveillance \\
013 & v1.9.1 & Opus 5.5 & unchanged (research model changed) & 0 & 0 & 0 & 0 & 0 & \textbf{15} & Machine translation of Japan's new weather-warning names \\
 
\bottomrule
\end{tabularx}
\end{table}

Formulating the research question starts from where to look, and this part of the instructions is the one we changed most. The system was developed over 13 runs with Japanese instructions; between runs we changed only the instructions and recorded the reason for each change. Table~\ref{tab:dev} lists, for each run, what the instructions said about the research area, the candidates the agent wrote, and the study it chose. These are observations, not controlled comparisons: the research model, its environment and other parts of the instructions also changed, and the classification of the candidates is ours.

\paragraph{No area.} The first run had no field and no selection procedure. The agent listed two candidates, chose an exact comparison in a classical probability problem, and finished in about 12 minutes of its three-hour limit.

\paragraph{Computer science.} From run 002 to run 010 the instructions set the field to computer science. The chosen studies stayed inside computer systems and AI tooling: caches (runs 002 and 010), the edit tools of coding agents (004), structured-generation engines (005) and document retrieval (009). In run 010, 7 of the 15 candidates concerned compression, storage, caching or deduplication, which are the kinds of constraint the instructions imposed on the run itself; the agent appeared to derive its topics from its own working conditions.

\paragraph{Single sentences and the agenda.} Within computer science, changes to single sentences coincided with changes in what the candidates were about. When the instructions welcomed ``problems of the AI era'' and scored candidates on that axis (v1.2), most candidates concerned AI or AI tooling for both Claude Opus~5.5 (9 of 16) and GPT-6 (8 of 11), and both runs chose AI tooling. When the readers were set to be future autonomous research agents (v1.3), Opus wrote mostly candidates about the behaviour of LLMs and about research practice (16 of 19 and 15 of 18 in runs 006 and 007) and chose one of them. When the readers were changed to researchers and practitioners of the field (v1.6), the next run of gpt-6.1-sol wrote no candidate about AI or research practice (the previous run of the same model wrote four) and 11 about computer systems. Conversely, when the area setting and the selection procedure were left almost unchanged (v1.7 mainly changed how to write the paper and added a search for classic prior work), run 010 repeated many of run 009's candidates (seven pairs by our matching), although neither run could see the other.

\paragraph{Societal problems.} In v1.8 we removed the field and added one sentence: ``The research field is not restricted. Freely devise research on themes such as discovering societal problems using open data and other information on the Internet, or proposing tools that help solve societal problems.'' The version also told the agent not to derive topics from the run's resource limits and named as readers those who work on the problem. With the same model and environment as runs 009 and 010, 14 of 15 candidates concerned societal problems (disasters and water, environment, energy, food, health and official statistics), none corresponded to a candidate of the two previous runs, and the chosen study moved from workflow caches to the monitoring of PFAS in drinking water. This was not more freedom but a different area: we did not test removing the field without adding the hint.

\paragraph{Technology $\times$ societal problem.} The candidates were still written from memory, and with the same model and procedure they repeated. In v1.9 we therefore made the divergence start from technologies: list at least six technologies of computer science that came into practical use in 2024 or later, generalise each into capabilities, and look for fields outside computer science where someone could not make a judgment because of the constraint that the capability removes (Section~\ref{sec:system}). Combined with the hint of v1.8, this sets the area that the evaluation runs used: a new technology applied to a societal problem. Both research models then wrote almost only societal candidates (15 of 17 with gpt-6.1-sol in run 012, 15 of 15 with Claude Opus~5.5 in run 013). The model changed the country and the subject rather than the area: 14 of the 15 candidates of run 013 used Japanese institutions and data, while the gpt-6.1-sol runs used none. Both chosen studies, however, used a known kind of technology (machine translation in run 013) or an existing application (a time-series forecaster already used for wastewater surveillance in run 012). The English version used in the evaluation therefore restricts the starting technologies to capabilities that first became possible around 2024 or later, excluding new versions of an existing kind of technology, and checks whether the technology has already been applied to the same problem.

\paragraph{The checks.} The six checks came with the technology-first divergence. The check that the main result does not follow from definitions or from how the data are produced answers the most frequent leading finding of the development reviews. The checks for decoration and for public ground truth guard against using a new technology without need or without a way to verify its judgments. In six short trials of the procedure that stopped once the candidates were written (not counted as runs), the new divergence lowered the rated feasibility of the candidates, so we added the checks that the technology or its published outputs can be used in the run's environment and that each possible answer changes a decision of the people who use the result. For the evaluation we translated the instructions into English, split the question and review procedures into skills, and froze them.

\paragraph{What this suggests.} Changes to the area setting coincided with the largest shifts in the candidates that we observed, and with entirely different chosen papers; the rest of the procedure acted within the area, on how well the question was formed. Their effects were not isolated from the other changes made at the same time. Because a one-sentence hint coincided with a shift of 14 of 15 candidates, we expect that a more specific setting, such as a particular field, problem, dataset or the open decision of a particular user, would let the same procedure produce correspondingly specific research themes while still leaving the question itself to the agent. We have not tested this. The evaluation runs also show the price of a fixed setting: within technology $\times$ societal problems, the agents converged on a few well-documented resources (Section~\ref{sec:res-questions}).

\section{Evaluation design}
\label{sec:design}

We first ran a pilot (run 014) with the English instructions and, after seeing its result, froze the system without changing it. We then pre-registered the evaluation: the system (SHA-256 of every file), the models and settings, the time and resource limits, the inclusion rules (every run whose start message is sent is reported; no run is excluded or repeated after its result is seen), and five counted runs, 015 to 019, excluding the pilot. After runs 015--017 had finished and their results had been seen, and before run 018 started, we amended the plan to count the pilot as the first run, because its filled instructions, skills, models and settings were identical to those of the later runs (verified by re-deploying the frozen files and comparing them); run 019 was therefore not made. The amendment added a run whose result had been seen and ended the series one run early, so we report it as a deviation. The pilot also differed in that the research workspace contained no earlier runs. Versions v1.10 to v1.10.4 differ only in the version label and the run number.

We report, for each run, the times, stops and inputs from the run records and the management records; the candidates and their fates; the critique and the agent's decisions on each point; and the in-loop review and the agent's response. All numbers in the tables are generated by a script from these records.

\paragraph{External AI evaluation (E1).} The pre-registration required re-review by AI reviewers from at least two model families other than the researcher's. Before sending any review, we fixed the prompt, the output schema and the scripts. Two reviewers, Gemini~4~Argon (thinking level high, Google Search tool) and Muse Wonder (reasoning effort max, web-search tool), each reviewed each final paper twice in independent requests. They saw only the paper and could search the web; they were not given the code, data or records. They were told that the paper was produced by an autonomous AI research system. They scored the six axes of our review instructions, gave a verdict, an overall score on the ICLR scale (1--10) as a full paper at a strong venue in the paper's own field, judgments of whether the paper would be accepted at a workshop in its field (acceptance rate about 70\%) and at the main track of a top venue in its field (about 25--30\%), and rated the research question. These are uncalibrated AI assessments of the papers alone, not calibrated against human decisions, and the 20 reviews are repeated assessments of five papers. The five papers are published as PDFs; Gemini first reviewed the Markdown sources, and after that we decided to review the PDFs, which are the published form, with both reviewers. We report the PDF reviews as the main result and the Markdown reviews in Appendix~\ref{app:e1}; this choice was made after the Markdown results had been seen.

\section{Results}
\label{sec:results}

\subsection{Completion and inputs}
\label{sec:res-ops}

\begin{table}[t]
\centering
\caption{Time and inputs. ``Topic'' is minutes from start to topic selection (excluding stops); ``Draft'', ``Stops'' and ``Active'' are hours:minutes from start to the draft tag, of usage-limit stops, and their difference; ``Margin'' is the time left before the draft deadline (12 h plus stops). ``Resumes'' counts fixed resumption messages after usage limits + after other stops. ``Perm.'' are permission requests and ``Advice'' research-content inputs recorded in the run's intervention log. ``Review'' and ``Resp.'' are minutes of the AI review and of the agent's response. ``API'' is the cost of the critique in US dollars, as recorded by the agent; in run 017 the agent recorded a range because it could not determine how long cached input was retained. Times are rounded from seconds, so Active may differ from Draft minus Stops by one minute.}
\label{tab:ops}
\small
\setlength{\tabcolsep}{4pt}
\begin{tabular}{@{}lrrrrrcccrrl@{}}
\toprule
Run & Topic & Draft & Stops & Active & Margin & Resumes & Perm. & Advice & Review & Resp. & API \\
\midrule
014 & 64 & 11:05 & 1:53 & 9:12 & 2:48 & 2 + 1 & 0 & 0 & 28 & 57 & 2.02 \\
015 & 54 & 14:30 & 2:48 & 11:42 & 0:18 & 3 + 1 & 0 & 0 & 44 & 49 & 2.02 \\
016 & 72 & 11:46 & 1:53 & 9:54 & 2:06 & 2 + 0 & 0 & 0 & 20 & 69 & 1.94 \\
017 & 33 & 7:26 & 0:59 & 6:27 & 5:33 & 1 + 0 & 0 & 0 & 31 & 42 & 1.95--2.02 \\
018 & 60 & 18:01 & 6:18 & 11:43 & 0:17 & 4 + 0 & 0 & 0 & 73 & 75 & 1.99 \\
 
\bottomrule
\end{tabular}
\end{table}

All five runs produced a first draft within the time limit, completed the AI review, and submitted a revised final version within the response period (Table~\ref{tab:ops}). The intervention logs record no research advice and no permission requests. After each start message, the research session received only fixed messages from the management agent: 12 resumptions after usage limits, 2 after other stops (the model was temporarily at capacity) and 5 review relays. The human set the API key and opened the terminal and took no other action in the runs. The agents worked for 6.5 to 11.7 active hours before the draft; usage-limit stops added up to 6.3 hours of waiting in one run. Each run made exactly one critique call, costing about US\$2.

The records were not free of errors. In three runs, after the agent's conversation was automatically compacted, the agent took a resumption message it had already answered for a new one and extended its own deadline by up to 260 seconds; in run 018 it noticed and withdrew the extra entry itself. We use the management records for all times.

\subsection{Questions the agents formulated}
\label{sec:res-questions}

\begin{table}[t]
\centering
\caption{The five runs and their questions (abridged from the paper titles). LCMAP: Land Change Monitoring, Assessment, and Projection; NLCD: National Land Cover Database; ECMWF: European Centre for Medium-Range Weather Forecasts.}
\label{tab:runs}
\small
\begin{tabularx}{\textwidth}{@{}llYY@{}}
\toprule
Run & Version & Question & Main data \\
\midrule
014 & v1.10 & Screening one-year tree-cover change in the contiguous US with annual satellite embeddings & LCMAP, NLCD; AlphaEarth, TESSERA embeddings \\
015 & v1.10.1 & Bounds on joint temperature-exceedance probabilities from published ensemble summaries & ECMWF ensemble forecasts \\
016 & v1.10.2 & Re-sampling reference locations to estimate change in Swiss crop-group diversity & Swiss crop declarations; annual embeddings \\
017 & v1.10.3 & Training-point allocation and recovery of small wetland patches & US National Wetlands Inventory; AlphaEarth, Sentinel-2 \\
018 & v1.10.4 & Do satellite embedding distances improve matching of wetland water regimes? & US National Wetlands Inventory; AlphaEarth, TESSERA; terrain and surface water \\
 
\bottomrule
\end{tabularx}
\end{table}

\begin{table}[t]
\centering
\caption{Candidate questions and the critique. ``Cand.'' is the number of candidates (3 obvious + 12--14 from the divergence), ``Tech.'' the number of technologies listed in the divergence; ``Points'' the critic's points on the top five, ``Adopted'' those adopted or partly adopted by the agent, and ``On chosen'' the points on the candidate that was finally chosen. In no run did the critique change which candidate was chosen.}
\label{tab:selection}
\small
\begin{tabular}{@{}lrrrrr@{}}
\toprule
Run & Cand. & Tech. & Points & Adopted & On chosen \\
\midrule
014 & 15 & 6 & 26 & 22 & 9 \\
015 & 15 & 6 & 26 & 21 & 7 \\
016 & 15 & 6 & 23 & 13 & 7 \\
017 & 15 & 6 & 23 & 19 & 6 \\
018 & 17 & 7 & 21 & 14 & 6 \\
 
\bottomrule
\end{tabular}
\end{table}

Table~\ref{tab:runs} lists the questions. All five are societal or environmental questions answered with public data and published model outputs: whether annual satellite embeddings can screen tree-cover change, how much published ensemble summaries constrain the probability that several places exceed a temperature threshold together, how to re-sample reference locations to estimate change in crop diversity, how the allocation of training points affects the recovery of small wetlands in a map, and whether embedding distances add to physical covariates when matching wetlands with similar water regimes. None of the five is about AI itself; four use annual satellite embeddings (AlphaEarth Foundations \citep{brown2025alphaearth} and TESSERA \citep{feng2025tessera}), which became available in 2025, as one of the objects of study rather than as decoration, and one uses ensemble weather forecasts. Run 015 is a boundary case of the requirement that a question depend on a newly possible capability: its selection record states that the mathematical question could have been studied before 2024 and that the new capability (cheap ensemble forecasts from AI weather models) supplies the motivation and a second class of real forecasts. The agent kept the candidate, as its own judgment of the checks allowed.

The agents wrote 77 candidate questions (Table~\ref{tab:selection}; counted per run, and similar candidates recur across runs). By our classification, with one field per candidate, all 77 concern society or the environment, in nine fields: environment (17), disaster and water (14), food and agriculture (12), health (7), transport and welfare (7), resources and materials (7), public administration (6), energy (4) and cultural heritage (3)%
 . None concerns AI or research practice, and none uses Japanese data. Topic selection took 33 to 72 minutes.

Most candidates were discarded for reasons that the six checks target: the technology or its outputs could not be used here (for example, the archive of a weather model's forecasts required an access approval, and the weights of a sound-separation model were gated), there was no public ground truth for the decision of interest, the result followed from a threshold definition, or close prior work already covered it. In run 018, the second-highest-scoring candidate (whether general sound separation preserves faint bird calls) was dropped because its model weights and outputs were not accessible.

The divergence repeatedly reached the same technologies and topics, although no run could see another. AlphaEarth satellite embeddings were listed in all five runs and used in four of the chosen studies; a machine-learning weather model (GenCast) was listed in all five, a time-series foundation model (Chronos) in four, and AlphaFold~3 in four. Each run's first ``obvious'' candidate was about heat, and the last two chosen studies both used the US National Wetlands Inventory. Within the procedure, the requirement to use a recent technology with public outputs and public ground truth narrows the field to a few well-documented resources.

\subsection{What the critique changed}
\label{sec:res-critic}

The critic raised 21 to 26 points per run, mostly adopted (Table~\ref{tab:selection}). It never changed the chosen candidate, but in every run it changed how the chosen question was compared or measured:
\begin{itemize}[leftmargin=*,itemsep=1pt]
  \item Run 014: the evaluation moved from the smoothness of embedding trajectories to held-out false alarms and misses, with calibration in earlier years and testing in later years, and gains were reported separately because they were rare.
  \item Run 015: the decision served by the published summaries was made explicit, and the exact distribution of the number of exceeding sites became the comparison; the critic pointed out, and the agent accepted, that the narrowing of nested bounds is mathematics, not a finding.
  \item Run 016: the reference data were changed from a classified map to farmers' declarations, because calibrating and validating with the same classified map would share its errors; uncertainty and abstention became primary endpoints.
  \item Run 017: inventory boundaries were treated as near-answer information, an image-only allocation was added as a control, and the endpoint was restricted to the recovery of inventory patches.
  \item Run 018: the central comparison became the increment of embedding distances beyond a strong baseline of public terrain, surface-water and drainage covariates; the question was framed as a label-masked diagnostic of agreement with inventory classes, not of field hydrology.
\end{itemize}
The agents verified the prior work that the critic cited before using it and recorded the points they did not adopt, typically new references they had not checked in full text or proposals for candidates they had rejected.

\subsection{In-loop review and revision}
\label{sec:res-review}

\begin{table}[t]
\centering
\caption{In-loop review (Claude Opus~5.5 with the run records) of the first drafts: verdict, scores (1--5) on correctness, verifiability, novelty, generality, usefulness and clarity, findings by severity, the agent's response (adopted + partly adopted), and claims in the final paper (exploratory in parentheses).}
\label{tab:review}
\small
\setlength{\tabcolsep}{4pt}
\begin{tabular}{@{}llcccccccccll@{}}
\toprule
Run & Verdict & Corr. & Verif. & Nov. & Gen. & Use. & Clar. & Crit. & Mod. & Minor & Response & Claims \\
\midrule
014 & major & 4 & 5 & 3 & 2 & 3 & 2 & 0 & 6 & 2 & 5 + 3 & 22 (14) \\
015 & minor & 4 & 5 & 2 & 2 & 3 & 2 & 0 & 4 & 3 & 5 + 2 & 20 (12) \\
016 & major & 4 & 5 & 2 & 2 & 3 & 2 & 0 & 6 & 3 & 5 + 4 & 13 (8) \\
017 & major & 3 & 5 & 3 & 2 & 3 & 2 & 0 & 4 & 2 & 6 + 0 & 11 (7) \\
018 & major & 3 & 5 & 3 & 2 & 3 & 2 & 1 & 5 & 4 & 8 + 2 & 17 (13) \\
 
\bottomrule
\end{tabular}
\end{table}

The reviewer re-ran every first draft from its single entry point and reproduced the reported numbers (Table~\ref{tab:review}); verifiability was 5 in all runs. In run 018, for example, the regenerated paper and claims file were byte-identical to the submitted ones, and refits of the primary and six control models reproduced the archived predictions exactly. Four drafts received ``major revision'' and one ``minor revision''. Novelty (2--3), generality (2) and clarity (2) were low in all runs.

The leading findings concerned interpretation, not execution. In run 014, the adjustment being tested had little to correct, so its null effect was largely predictable. In run 015, much of the result followed from exact moment properties of a 50-member ensemble. In run 016, the same declaration database served as both truth and audit labels. In run 017, the abstract led with the largest of four split estimates and omitted planned conditions with different results. In run 018, the only critical finding of the series: planned null controls (permuted embeddings, other random seeds) produced changes as large as the real embeddings, yet the draft's abstract presented the increment without applying the qualification rule the agent had itself written before seeing the results. All 40 findings were adopted (29) or partly adopted (11). The agents rewrote abstracts around the answer, added the planned conditions they had omitted, and in run 018 restated the main result as ``an increment attributable to the embeddings was not established''. The final versions were not reviewed again.

\subsection{What the final papers report}
\label{sec:res-answers}

\begin{table}[t]
\centering
\caption{What each final paper reports, summarised from its abstract: the answer to its question and the main limitation the paper itself states. Numbers are left out; they are in the papers.}
\label{tab:answers}
\small
\begin{tabularx}{\textwidth}{@{}lYY@{}}
\toprule
Run & Answer in the final paper & Main limitation stated by the paper \\
\midrule
014 & Reference and control corrections of annual-embedding change scores did not consistently improve the screening of one-year tree-cover gains and losses; a directly trained classifier scored higher, with unequal label resources. & Small one-year tests; exploratory changes of the data partition reversed all correction contrasts; performance on the native grid is unverified; multi-year restoration and ecological recovery are not addressed. \\
015 & Published single-site exceedance probabilities leave wide bounds on the probability that at least $k$ of a set of sites exceed a threshold; one aggregate moment narrows them, and all pairwise co-exceedance probabilities nearly determine it. & The near-identification is conditional on exact moments of the 50-member empirical forecast law; no observed-weather skill, health benefit or realised economic value was measured. \\
016 & Revisiting the same reference locations gave much smaller errors in the annual change of declared crop-group diversity than independent locations, but predicting no change was better still at every budget; assistance from satellite embeddings gave no or modest gains. & Declaration labels are treated as perfectly repeatable on a fixed frame; field-audit errors, costs and ecological outcomes were not measured. \\
017 & With patch-first allocation of training points, annual embeddings recovered more small inventory wetland patches than seasonal optical imagery in the planned split, but the gain did not transfer reliably between projects. & Intervals exclude the uncertainty of split selection; the endpoint is agreement with the mapped inventory, not ecological completeness or savings in annotation cost. \\
018 & An increment of embedding distances over terrain, surface-water and drainage covariates for matching wetland water regimes was not established: planned permutation controls gave changes comparable to the real distances. & Convenience windows; the endpoint is agreement with inventory labels, not field hydroperiod or benefit to practitioners. \\
\bottomrule
\end{tabularx}
\end{table}

Table~\ref{tab:answers} summarises the answers. Two of the five main results are negative or null (runs 014 and 018). In run 016, revisiting reduced the error, but predicting no change was better still. Runs 015 and 017 report positive findings with stated limits on transfer and on what the endpoint measures. Each paper states that its endpoint is agreement with published labels or with a forecast distribution, not the field outcome that motivates it.

\subsection{External AI evaluation}
\label{sec:res-e1}

\begin{table}[t]
\centering
\caption{External AI evaluation of the final papers (PDF, paper and web search only, two reviews per paper). Axis scores are means over the two reviews; ``Overall'' lists the two ICLR-scale scores; verdicts are m (minor revision), M (major revision) or R (reject); WS and Main are workshop-level and main-track-level judgments, Y (yes), B (borderline) or N (no).}
\label{tab:e1}
\footnotesize
\setlength{\tabcolsep}{4pt}
\begin{tabular}{@{}llccccccllll@{}}
\toprule
Reviewer & Run & Corr. & Verif. & Nov. & Gen. & Use. & Clar. & Overall & Verdict & WS & Main \\
\midrule
Gemini 4 Argon & 014 & 4.0 & 4.0 & 3.0 & 2.0 & 3.0 & 2.0 & 5/5 & M/M & Y/Y & N/N \\
 & 015 & 5.0 & 5.0 & 3.0 & 3.0 & 3.0 & 3.0 & 6/6 & m/m & Y/Y & B/B \\
 & 016 & 4.5 & 5.0 & 3.0 & 2.5 & 3.5 & 2.0 & 6/5 & M/M & Y/Y & B/N \\
 & 017 & 4.0 & 4.0 & 3.0 & 3.0 & 3.0 & 2.5 & 6/5 & m/M & Y/Y & B/N \\
 & 018 & 4.5 & 3.5 & 3.0 & 2.5 & 3.0 & 2.0 & 5/5 & M/M & Y/Y & N/N \\
& \emph{mean} & 4.4 & 4.3 & 3.0 & 2.6 & 3.1 & 2.3 & 5.4 & & & \\\midrule
Muse Wonder & 014 & 3.5 & 3.0 & 2.0 & 2.0 & 3.0 & 2.0 & 4/3 & M/R & Y/B & N/N \\
 & 015 & 4.0 & 3.0 & 2.0 & 2.0 & 3.0 & 2.0 & 4/4 & R/M & B/B & N/N \\
 & 016 & 4.0 & 2.5 & 2.0 & 2.0 & 3.0 & 2.0 & 4/5 & M/M & B/B & N/N \\
 & 017 & 3.5 & 3.0 & 2.0 & 2.0 & 3.0 & 2.0 & 5/4 & M/M & Y/B & N/N \\
 & 018 & 4.0 & 2.5 & 3.0 & 2.0 & 3.0 & 2.0 & 4/4 & M/M & B/Y & N/N \\
& \emph{mean} & 3.8 & 2.8 & 2.2 & 2.0 & 3.0 & 2.0 & 4.1 & & & \\
 
\bottomrule
\end{tabular}
\end{table}

All 20 reviews returned valid output; none reported recognising the papers. Gemini~4~Argon judged all five papers workshop level, with overall scores of 5 or 6 (mean 5.4) and main-track judgments of ``borderline'' or ``no''. Muse Wonder was stricter: overall 3 to 5 (mean 4.1), workshop level ``yes'' in 3 and ``borderline'' in 7 of 10 reviews, main track ``no'' in all, and two ``reject'' verdicts (runs 014 and 015). The two reviewers disagreed most on run 015 (minor revision and 6/10 versus reject or major revision and 4/10). They agreed on the profile: correctness high (4.4 and 3.8), novelty (3.0 and 2.2), generality (2.6 and 2.0) and clarity (2.3 and 2.0) low, the same profile as the in-loop reviews. They rated the research questions themselves at 3.2 and 3.4 out of 5, and in 19 of 20 reviews they judged the main answer ``partly'' predictable from the design. Verifiability from the paper alone was rated 4.3 by Gemini and 2.8 by Muse, below the in-loop 5.0; the in-loop reviewer had the records, but it was also a different model and reviewed the first drafts. The two reviews of the same paper differed by 0.4 (Gemini) and 0.6 (Muse) points overall on average.

\section{Discussion}
\label{sec:discussion}

\paragraph{The question can be left to the AI.} With a plain agent, an instruction document and two skills, five consecutive runs formulated their own questions and carried them through to revised, re-computable papers without research input from people. The human set only the area, in one or two sentences; neither the specific question nor the outputs to report were chosen by people, which makes the AI's choices observable.

\paragraph{What the AI chose, and what the instructions chose.} Within the technology $\times$ societal-problem setting, the agents chose societal and environmental questions on public data, and the technology-first procedure pulled them toward a small set of well-documented resources. The area itself came from the instructions: in the development runs, the same research model chose caches and document retrieval when the field was computer science, and societal problems after one sentence was changed (Section~\ref{sec:area}). Within the area, the convergence is partly a property of the procedure (a recent technology with public outputs and public ground truth) and partly of the model; we did not vary the model in this series. It suggests that an AI researcher's agenda, once the area is set, concentrates where data and outputs are open. Leaving the question to the AI therefore does not remove the human choice of area; it makes that choice small, explicit and testable.

\paragraph{Critique and review do different jobs.} The critic, consulted after short feasibility probes but before the main plan, shaped what was compared and what was claimed; the reviewer, with the records, checked that the claims followed from the evidence and that planned analyses were reported. Novelty and generality of the final papers remained low in both external reviews; since we did not evaluate the papers before and after the critique and the review, we cannot estimate how either affected them. The weakest point of the system remains the choice of a question whose answer matters and is not fixed by the design.

\paragraph{Execution did not limit these runs.} Execution did not prevent completion in these five runs, which chose their questions partly for feasibility. A separate reviewer agent re-ran every first draft, and every finding was answered within the response period. The remaining problems, as in our development runs, were in framing: which estimate leads the abstract, whether controls are given their planned weight, and whether the answer was predictable.

\paragraph{Future work.} The area setting deserves its own experiments. All five evaluation runs used one setting, a new technology applied to a societal problem. Settings other than technology $\times$ societal problems remain to be tested, for example a named scientific field, a given dataset, theoretical questions, or the open decision of a particular user, as does a setting with no hint at all, which would separate the effect of removing the field from that of adding the hint. A more specific setting may also yield more specific themes, and whether the procedure then still produces non-obvious questions is open. A human evaluation of the five papers is planned separately.

\section{Limitations}
\label{sec:limitations}
Five runs with one research model are a small sample, and we made no ablation, so we cannot attribute outcomes to the question procedure, the critic or the reviewer. All evaluations are by AI models; a human evaluation is planned separately. The in-loop reviewer is from the same family as the management agent. We counted the pilot run after seeing three results, and we chose the PDF reviews as the main external result after seeing the Markdown reviews; both are reported as deviations. The external reviewers read only the papers, so their verifiability scores do not reflect the records. Topic selection consulted a critique by another model, and the hint toward societal problems in the instructions shaped the domain. The lessons on the area setting (Section~\ref{sec:area}) come from development runs in which the research model, its environment and other parts of the instructions also changed, so they are observations rather than controlled comparisons. Re-computation by the reviewer re-ran the agent's own code on the first draft; it is not an independent implementation, and it does not establish the correctness of reference labels or the validity of interpretations. Isolation from earlier runs relied on separate configuration and on the instructions, not on access control. The external ratings are uncalibrated AI judgments of the papers alone. This paper was drafted by the management agent from the records (Appendix~\ref{app:declaration}).

\section{Conclusion}
A coding agent given an instruction document and two skills can formulate its own research question and carry it through to a revised, verifiable paper, repeatedly and without research input from people. AI reviewers reading only the papers judged them favourably or borderline for a workshop in their fields and below the main track, without calibration against human decisions, and rated their novelty and generality low. In our development runs, what kind of research resulted shifted with the one or two sentences that set the research area, and within that area depended on the procedure that forms the question. Formulating questions that matter is where autonomous research agents now need the most work, and leaving the question to them is how to study it.

\paragraph{Availability.} The system (instructions, skills, operating procedure), the five papers as PDFs with their Markdown sources, the records needed to reproduce them (plans, events, claims, reviews and responses, code; external data as retrieval procedures and checksums), and the evaluation scripts are available at \url{https://github.com/kofujimura/autonomous-research-agents} (code under the MIT License; text and records under CC~BY~4.0; third-party data under their original terms). The five papers are attributed to ``Autonomous Research Agents''; their wording is unchanged from the final versions.

\bibliographystyle{plainnat}
\bibliography{refs}

\appendix

\section{External AI evaluation of the Markdown sources}
\label{app:e1}
Before the papers were converted to PDF, Gemini~4~Argon reviewed their Markdown sources with the figures attached as images (same prompt except for one sentence on the input form; two reviews per paper). The ten reviews gave mean scores of 4.4 (correctness), 4.3 (verifiability), 2.9 (novelty), 2.7 (generality), 3.4 (usefulness) and 2.4 (clarity), and an overall score of 5.4, close to the PDF reviews. The prompts, schema, scripts, raw responses and costs are in the repository.

\section{Statement on AI involvement}
\label{app:declaration}
The five studies were conducted and written by AI agents as described. This paper was drafted by Claude Opus~5.5, acting as the management agent of the project, from the project records at the request of the author, who decided the scope and claims, reviewed the full text and directed its revisions. The numbers in the tables are generated by scripts from the records.

\end{document}